\documentclass{article}

\usepackage[preprint]{neurips_2026}
\usepackage{url}            
\usepackage{booktabs}       
\usepackage{amsfonts}       
\usepackage{microtype}      
\usepackage{xcolor}         
\usepackage{tabularx}
\usepackage{float}
\usepackage{hyperref}
\usepackage{makecell}
\usepackage{graphicx}
\usepackage{amsmath}
\usepackage{enumitem}

\definecolor{boxborder}{RGB}{45,55,180}
\definecolor{boxbg}{RGB}{245,245,250}
\definecolor{reflink}{RGB}{45,55,220}

\definecolor{mygreen}{RGB}{0,150,136}
\definecolor{myred}{RGB}{220,60,80}
\definecolor{hlpink}{RGB}{245,230,235} 

\definecolor{myteal}{RGB}{0,92,92}
\definecolor{mygraybg}{RGB}{245,245,245}

\newcommand{\cmark}{\textcolor{mygreen}{$\checkmark$}}
\newcommand{\xmark}{\textcolor{myred}{$\times$}}
\newcommand{\xxmark}{\textcolor{myred}{$\times$}}

\title{LongBench: Evaluating Robotic Manipulation Policies on Real-World Long-Horizon Tasks}

\author{%
\begin{tabular}{c@{\hspace{1.2em}}c@{\hspace{1.2em}}c@{\hspace{1.2em}}c@{\hspace{1.2em}}c@{\hspace{1.2em}}c}
Xueyao Chen & Jingkai Jia & Tong Yang & Yibo Fu & Wei Li & Wenqiang Zhang
\end{tabular}\\[0.5em]
Fudan University \\[0.35em]
\texttt{\{xueyaochen24, jkjia24, tongyang23, ybfu22\}@m.fudan.edu.cn} \\[0.2em]
\texttt{\{fd\_liwei, wqzhang\}@fudan.edu.cn}
}

\begin{document}

\maketitle

\begin{abstract}
Robotic manipulation policies often degrade over extended horizons, yet existing benchmarks provide limited insight into why such failures occur. 
Most prior benchmarks are either simulation-based or report aggregate success, making it difficult to disentangle the distinct sources of temporal difficulty in real-world execution.
We introduce LongBench, a real-world benchmark for evaluating long-horizon manipulation. 
LongBench consists of over 1,000 real-world episodes, covering two complementary regimes: Context-Independent (fully observable) and Context-Dependent (ambiguity-driven). 
By organizing tasks into capability- and ambiguity-specific subsets, LongBench enables mechanism-aware evaluation of execution robustness, temporal consistency, and context-dependent reasoning.
Evaluating six state-of-the-art policies reveals that long-horizon performance is not governed by a single factor. 
We observe that performance in fully observable settings is more strongly associated with execution robustness, while contextual difficulty varies across tasks and is not consistently improved by memory-based methods. 
We hope that LongBench serves as a useful benchmark for studying long-horizon manipulation and for developing policies with stronger robustness across both execution and contextual challenges. Policy submission and leader board page: \url{https://cxy0103.github.io/LongBench_Website/}.
\end{abstract}
\section{Introduction}
\label{sec:intro}

Robotic manipulation is a central capability of embodied intelligence, requiring agents to act reliably over extended time horizons in dynamic physical environments. While recent advances in vision–language–action (VLA) and vision–action (VA) policies have substantially improved short-horizon skill execution, robust long-horizon manipulation remains challenging. In real-world settings, errors accumulate over time, perceptual noise interferes with state estimation, and sub-tasks are tightly interdependent, so small early deviations may invalidate later stages. As a result, evaluating policies solely on short or isolated skills fails to capture the temporal robustness required for sustained physical interaction.

Most existing long-horizon benchmarks are developed in simulation~\citep{rlbench, calvin, libero, vlabench, robocas, robocerebra}, while real-world datasets such as furniture assembly and multi-task demonstration collections~\citep{heo2023furniturebench} provide complex multi-stage behaviors under relatively structured conditions. Although these efforts have advanced long-horizon evaluation, they typically treat long-horizon manipulation as a single difficulty axis, relying on aggregate success rates to measure performance. However, this view is incomplete: long-horizon tasks do not share a single underlying challenge, and aggregate metrics fail to distinguish between different failure sources, such as execution instability, dynamic response limitations, and contextual ambiguity.

To address this gap, we introduce LongBench, a real-world benchmark for mechanism-aware evaluation of long-horizon robotic manipulation. LongBench is designed not only to measure performance, but to expose the underlying structure of long-horizon behavior. It studies two complementary regimes of temporal difficulty: (i) \textbf{Context-Independent long-horizon tasks}, where all decision-relevant information is present in the current observation and difficulty arises from extended execution dynamics; and (ii) \textbf{Context-Dependent long-horizon tasks}, where observations are state-aliased and require policies to retain and bind earlier context across extended horizons.

Rather than treating long-horizon manipulation as a monolithic problem, LongBench decomposes it into capability- and ambiguity-specific subsets. This design enables fine-grained diagnosis of distinct failure modes, including execution robustness under prolonged interaction, dynamic response under short-lived action opportunities, and context-dependent reasoning under ambiguity. By grounding these evaluations in real-world execution, LongBench exposes interactions between perception, control, and temporal structure that are difficult to capture in simulation.

We provide a standardized real-world evaluation platform, including a testing API and environment replication protocols, enabling reproducible and comparable assessment of robotic manipulation policies. 
Using this framework, we evaluate six state-of-the-art policies, including both memory-based and frame-based architectures.  Our analysis reveals that long-horizon performance is not governed by a single factor, but by distinct and sometimes competing mechanisms. In particular, execution robustness dominates in fully observable settings, while contextual difficulty is heterogeneous and not uniformly addressed by memory-based methods. These findings highlight structural limitations of current policies and demonstrate the need for benchmarks that diagnose underlying failure mechanisms.

Our contributions are fourfold:
\begin{itemize}[leftmargin=1.5em, itemsep=0.2em, topsep=0.2em]
\item A real-world long-horizon benchmark that separates two regimes of temporal difficulty, Context-Independent (fully observable) and Context-Dependent (ambiguity-driven), enabling mechanism-aware evaluation of robotic manipulation;
\item A dataset of 1,000 real-world demonstrations spanning diverse tasks and scene contexts;
\item A diagnostic evaluation protocol that decomposes long-horizon performance into capability- and ambiguity-specific subsets, providing a diagnostic lens to reveal distinct failure modes beyond aggregate success rates;
\item A comprehensive empirical study that uncovers structural properties of long-horizon behavior, including execution bottlenecks, heterogeneous contextual difficulty, and trade-offs between execution and context retention.
\end{itemize}

\begin{table}[t]
\centering
\small
\caption{
Comparison of long-horizon manipulation benchmarks along key dataset and task dimensions, including real-world evaluation, long-context reasoning, dynamic interactions, and teleoperation-based data collection.
}
\label{tab:lb_compare}
\resizebox{0.98\columnwidth}{!}{
\begin{tabular}{lcccc}
\toprule
\textbf{Benchmark} & \textbf{Real World} & \textbf{Long-Context} &
\textbf{Dynamic} & \textbf{Teleop. Data} \\
\midrule
RLBench \citep{rlbench} & \xxmark & \xxmark & \xxmark & \xxmark \\
CALVIN \citep{calvin}   & \xxmark & \xxmark & \xxmark & \cmark \\
Libero-Long \citep{libero} & \xxmark & \xxmark & \xxmark & \cmark \\
FurnitureBench \citep{heo2023furniturebench}  & \cmark & \xxmark & \xxmark & \cmark \\
VLABench \citep{vlabench}  & \xxmark & \cmark & \xxmark & \xxmark \\
RoboCasa \citep{robocas} & \xxmark & \cmark & \xxmark & \cmark \\
RoboCerebra  \citep{robocerebra}  & \xxmark & \cmark & \cmark & \cmark \\
LongBench (Ours) & \cmark & \cmark & \cmark & \cmark \\
\bottomrule
\end{tabular}
}
\end{table}


\section{Related Work}
\label{sec:formatting}

\paragraph{Simulation Benchmarks for Long-Horizon Manipulation.}
Simulation-based benchmarks have enabled scalable evaluation of multi-step manipulation and language-conditioned control. Early suites such as RLBench~\citep{rlbench} established multi-task and instruction-following evaluation on tabletop scenes, while CALVIN~\citep{calvin} and LIBERO~\citep{libero} extended horizon length, compositional task structure, and language grounding to study multi-step skill chaining and knowledge transfer. More recent VLA-oriented benchmarks, including VLABench~\citep{vlabench}, RoboCerebra~\citep{robocerebra}, and RoboCAS~\citep{robocas}, emphasize commonsense reasoning, long instruction sequences, and planning under complex spatial arrangements. These platforms substantially expand temporal depth and semantic diversity, providing controlled environments for studying large-scale visuomotor and vision–language–action policies. However, simulation settings typically abstract away persistent physical execution dynamics and simplify perceptual uncertainty. As a result, they rarely disentangle distinct sources of temporal failure—such as cumulative control drift, recovery robustness, or delayed ambiguity resolution—that arise in real-world long-horizon execution.

\noindent \textbf{Real-World Benchmarks for Long-Horizon Manipulation.}
Real-world robotic manipulation benchmarks have become increasingly common, with recent efforts focusing on standardized evaluation~\citep{khargonkar2024scenereplica, rgmc2025} and scalable infrastructure~\citep{atreya2025roboarena, zhou2025autoeval}. However, benchmarks that specifically target long-horizon manipulation in real-world settings remain relatively limited.
FurnitureBench~\citep{heo2023furniturebench} provides a canonical real-world setup for multi-step manipulation, enabling evaluation of extended task sequences under consistent hardware conditions.
However, existing real-world long-horizon benchmarks primarily focus on reproducibility and aggregate task success, and do not explicitly decompose long-horizon difficulty into distinct components. 
As a result, it remains unclear which aspects of long-horizon behavior, such as execution stability or context-dependent reasoning, limit policy performance.
In contrast, our benchmark explicitly decomposes long-horizon manipulation into capability- and ambiguity-specific subsets, enabling more fine-grained analysis of policy performance.

\noindent \textbf{Vision-Language-Action Models.}
Recent advances in visuomotor and vision-language-action (VLA) policies have significantly improved performance on multi-step manipulation tasks. 
Large-scale pretrained policies such as $\pi_0$~\citep{pi0} leverage flow-based or diffusion-based action representations, while models such as SmolVLA~\citep{smolvla} and OpenVLA-OFT~\citep{kim2025fine} explore efficient fine-tuning and lightweight architectures. 
Diffusion-based approaches~\citep{chi2023diffusionpolicy} model action generation as a denoising process, enabling flexible multi-step prediction. 

To address long-horizon dependencies, recent work introduces explicit temporal modeling mechanisms, including memory-augmented policies such as MemoryVLA~\citep{shi2026memoryvla} and multi-frame architectures such as CronusVLA~\citep{li2025cronusvla}. 
These methods aim to improve performance under partial observability and long-context settings by incorporating historical information.
However, despite these advances, it remains unclear how different policy designs behave under distinct sources of long-horizon difficulty.

\section{LongBench Benchmark}
\label{sec:LongBench}
Existing long-horizon manipulation benchmarks differ along several dataset- and setup-level axes, including real-world vs.\ simulation settings, long-context requirements, dynamic interactions, and data collection strategies. Table~\ref{tab:lb_compare} summarizes these differences.
However, these attributes alone provide limited insight into how long-horizon behavior is structured or where policies fail. To address this limitation, we design LongBench with the following principles:

\textbf{(1) Real-world long-horizon evaluation.}
Many existing benchmarks are primarily simulation-based, which simplifies physical dynamics and reduces execution variability. In contrast, LongBench focuses on real-world manipulation, where noise, contact dynamics, and perception errors play a critical role.

\textbf{(2) Explicit separation of execution and context.}
Prior work often treats long-horizon manipulation as a single difficulty axis. LongBench instead separates two regimes: Context-Independent tasks, which emphasize execution over extended horizons, and Context-Dependent tasks, which introduce ambiguity requiring historical context. This separation enables more fine-grained analysis of policies.

\textbf{(3) Mechanism-aware task decomposition.}
Rather than reporting aggregate success alone, LongBench organizes tasks into capability- and ambiguity-specific subsets. This design enables diagnosing failure modes across distinct mechanisms, rather than relying solely on aggregate success rates.

Together, these design choices distinguish LongBench from prior benchmarks by combining real-world evaluation with mechanism-aware diagnostic structure.


\begin{figure}[t]
  \centering
  \includegraphics[page=1,width=\textwidth]{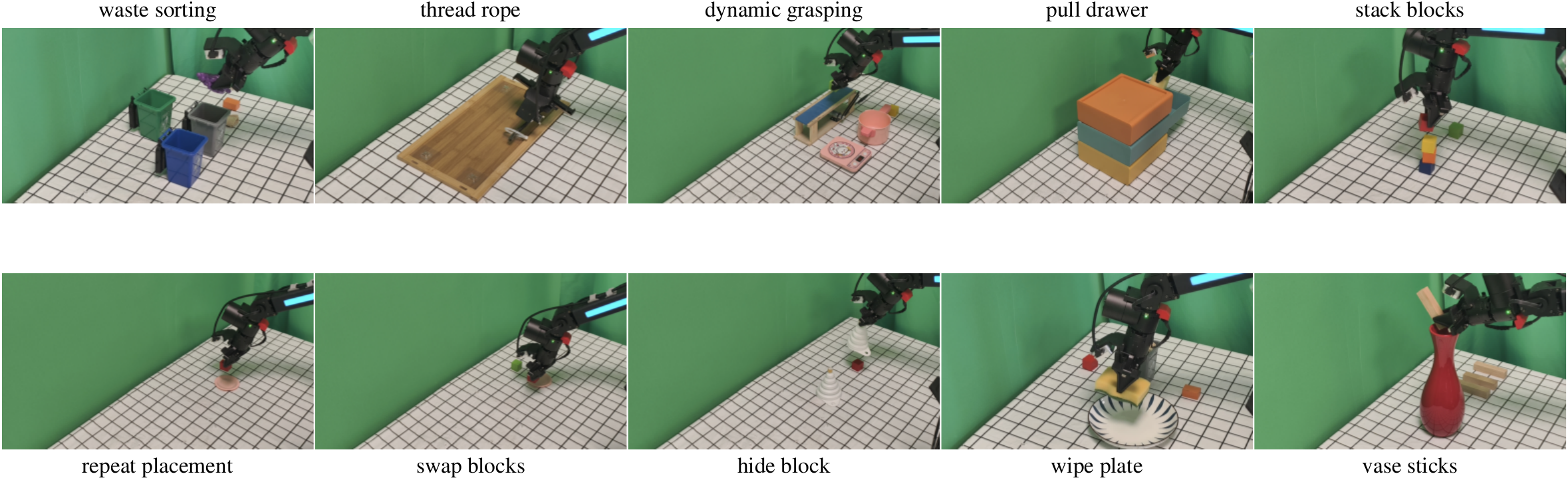}
  \caption{
  Overview of the 10 LongBench tasks, grouped into two regimes: Context-Independent (top row) and Context-Dependent (bottom row). 
  Context-Independent tasks emphasize sustained execution under full observability, while Context-Dependent tasks require resolving ambiguity using historical context. 
  Detailed task specifications are provided in Appendix.}
  \label{fig:task-dist}
\end{figure}

\subsection{Task Taxonomy}

Long-horizon manipulation is often treated as a single problem, where performance is measured by aggregating task success over extended horizons. 
However, tasks that are long do not necessarily share the same underlying challenges: some require sustained execution under full observability, while others depend on resolving ambiguity using historical context.

To capture this distinction, we organize LongBench into two complementary regimes, as illustrated in Figure~\ref{fig:task-dist}:

\textbf{Context-Independent Long-Horizon Tasks.}
In this regime, the next action can be determined from the current observation, including visual input, robot state, and language instruction. 
Difficulty arises from maintaining stable and feasible execution over long horizons, such as handling error accumulation, phase transitions, and time-critical interactions.

\textbf{Context-Dependent Long-Horizon Tasks.}
In this regime, the current observation is insufficient to determine the correct action. 
Visually similar states may correspond to different underlying conditions, requiring the policy to retain and use historical context to resolve ambiguity.

This separation is important because these two regimes expose fundamentally different failure modes. 
Execution failures in Context-Independent tasks arise from instability over long rollouts, while failures in Context-Dependent tasks stem from ambiguity that cannot be resolved without context. 
By separating them, LongBench enables more precise diagnosis of policy behavior, rather than conflating distinct challenges under a single performance metric.

\subsection{Context-Independent Long-Horizon Tasks}

We define Context-Independent long-horizon tasks as multi-step manipulation problems whose next-step decision can be made from the \emph{current} observation. 
The primary challenge in this regime is sustaining reliable execution over extended horizons.
Specifically, these tasks require maintaining a long chain of tightly-coupled perception--action loops, while preserving feasibility across stages. 
Failures typically arise from accumulated execution errors, incorrect phase transitions, or delayed responses under time constraints.

To characterize these challenges, we describe Context-Independent tasks using four \emph{task-structural} properties:

\textbf{(i) Phase Dependence.}
Tasks contain phase transitions with strict preconditions, where early mistakes can make later stages infeasible or require recovery. This introduces strong structural dependencies across task stages.

\textbf{(ii) Iterative Progress.}
Tasks involve sequences of similar subgoals over extended horizons, where each step produces observable progress toward completion. 
The challenge lies in correctly tracking task progress and advancing through these subgoals in the correct order.

\textbf{(iii) Error Accumulation.}
Over long rollouts, small execution errors, such as pose bias or timing inaccuracies, accumulate over time and gradually degrade performance. Unlike phase-dependent failures, these errors are often locally unnoticeable but compound across steps, eventually leading to global instability and failure.

\textbf{(iv) Temporal Windows.}
Some task stages require actions to be executed within limited time windows, where only timely decisions can lead to successful outcomes. Delayed responses may miss transient opportunities and make subsequent execution infeasible.

A compact per-task summary of phases, average episode length, and the dominant subset of these four properties is provided in Table~\ref{tab:ci_lh_props}.

\begin{table}[t]
\centering
\setlength{\tabcolsep}{3.5pt}
\renewcommand{\arraystretch}{1.05}
\caption{
Context-Independent Long-Horizon tasks organized by structural capability subsets. 
Each task is labeled according to the presence of four capability types: \textbf{PD} (Phase Dependence), \textbf{IP} (Iterative Progress), \textbf{EA} (Error Accumulation), and \textbf{TW} (Temporal Windows). 
Phases denote atomic sub-steps, and Avg.\ length refers to the average episode length in steps.
}
\label{tab:ci_lh_props}

\resizebox{0.75\linewidth}{!}{%
\begin{tabular}{lccccccc}
\toprule
\textbf{Task} & \textbf{Phases} & \textbf{Demos} & \textbf{Avg.\ length} &
\textbf{PD} & \textbf{IP} & \textbf{EA} & \textbf{TW} \\
\midrule
waste sorting     & 6  & 100 & 693 & \xmark & \cmark & \xmark & \xmark \\
thread rope       & 5  & 100 & 600 & \cmark & \cmark & \cmark & \xmark \\
pull drawer       & 8  & 100 & 960 & \cmark & \cmark & \xmark & \xmark \\
stack block       & 8  & 100 & 692 & \cmark & \cmark & \cmark & \xmark \\
dynamic grasping  & 6  & 100 & 481 & \cmark & \xmark & \xmark & \cmark \\
\bottomrule
\end{tabular}%
}
\end{table}

\subsection{Context-Dependent Long-Horizon Tasks}

We characterize Context-Dependent long-horizon tasks through ambiguity patterns that define how historical context is required for correct action prediction. 
Unlike Context-Independent tasks, the primary challenge in this regime is not sustained execution, but resolving ambiguity that cannot be disambiguated from the current observation alone. 
Failures typically arise when policies are unable to use earlier context to distinguish between visually similar but behaviorally different states.

To capture these challenges, we describe Context-Dependent tasks using four \emph{ambiguity patterns}:

\textbf{(i) Completion ambiguity.}
The current observation is indistinguishable from a valid terminal state, making it difficult to determine whether the task has been completed. 
This can lead to premature termination or incomplete execution.

\textbf{(ii) Count ambiguity.}
The current observation does not indicate how many times a repeated action has been executed, making it unclear whether to continue or stop. 
This can result in missing or redundant operations.

\textbf{(iii) Subtask-branch ambiguity.}
Visually similar observations may correspond to different subtask branches that require incompatible actions. 
Correct behavior depends on identifying which branch the agent is currently following.

\textbf{(iv) Cross-episode ambiguity.}
Similar observations may appear across different episodes with different task-specific requirements. 
Correct action selection therefore depends on distinguishing between these contexts.

A compact summary of task structure, phase decomposition, and ambiguity types is provided in Table~\ref{tab:cd_lh_props}.

\begin{table}[t]
\centering
\scriptsize
\setlength{\tabcolsep}{3.5pt}
\renewcommand{\arraystretch}{1.05}
\caption{
Context-Dependent Long-Horizon tasks organized by ambiguity patterns. 
Each task is labeled according to the presence of four ambiguity types: \textbf{CP} (Completion), \textbf{CT} (Count), \textbf{SB} (Subtask-Branch), and \textbf{CE} (Cross-Episode). 
Phases denote atomic sub-steps, and Avg.\ length refers to the average episode length in steps.
}
\label{tab:cd_lh_props}

\resizebox{0.75\linewidth}{!}{%
\begin{tabular}{lccccccc} 
\toprule
\textbf{Task} & \textbf{Phases} & \textbf{Demos} & \textbf{Avg.\ length} &
\textbf{CP} & \textbf{CT} & \textbf{SB} & \textbf{CE} \\
\midrule
repeat placement  & 8  & 100 & 773 & \cmark & \cmark & \xmark & \xmark \\
swap blocks       & 10 & 100 & 889 & \xmark & \xmark & \cmark & \xmark \\
wipe plate        & 8  & 100 & 651 & \xmark & \cmark & \xmark & \xmark \\
hide block        & 10 & 100 & 556 & \cmark & \xmark & \cmark & \cmark \\
vase sticks       & 6  & 100 & 510 & \cmark & \xmark & \xmark & \cmark \\
\bottomrule
\end{tabular}%
}
\end{table}
\section{Experimental Setup}
\label{sec:Experimental Setup}

\subsection{Platform and Data Collection}

LongBench is built on a tabletop manipulation platform centered on a 6-DoF ARX-R5 robotic arm. 
For visual observation, we use two RGB cameras: a fixed top-down camera providing a global view of the workspace and a wrist-mounted camera aligned with the end-effector for close-range perception. 
All image streams are recorded at $320\times240$ resolution and 20\,Hz. Robot actions and states are logged at the same frequency.

Demonstrations are collected via leader--follower teleoperation, where a leader arm directly controls the ARX-R5 as the follower. 
All visual streams, robot states, and actions are timestamped and synchronized. 
Benchmark evaluation uses the same platform and observation/control setup, ensuring consistency between data collection and policy evaluation.

\subsection{Evaluated Policies}

We evaluate representative policies across single-frame and history-conditioned settings.

\noindent\textbf{Single-frame policies.}
These models predict actions using only the current observation and do not explicitly incorporate temporal history.
\textit{$\pi_0$}~\citep{pi0}.
A pretrained VLA with a flow-matching continuous action head, representing large-scale pretrained Markovian policies.
\textit{OpenVLA-OFT}~\citep{kim2025fine}.
An efficiently fine-tuned VLA that performs parallel continuous action regression.
\textit{SmolVLA}~\citep{smolvla}.
A lightweight VLA trained from pretrained visual--language backbones with chunked action prediction.
\textit{Diffusion Policy}~\citep{chi2023diffusionpolicy}.
A conditional diffusion-based visuomotor model that generates multi-step action sequences.

\noindent\textbf{Multi-frame policies.}
These models explicitly incorporate past observations to capture temporal dependencies across time.
\noindent\textit{MemoryVLA}~\citep{shi2026memoryvla}.
A memory-augmented VLA that retrieves historical context from a compact memory bank during action generation.
\noindent\textit{CronusVLA}~\citep{li2025cronusvla}.
A multi-frame VLA that aggregates cross-frame features through a lightweight temporal decoder.

\subsection{Evaluation Protocol}
To enable fair comparison across methods, all policies are evaluated under a unified observation--action interface. 
Unless otherwise specified, policies receive synchronized top and wrist RGB observations at $320\times240$ resolution and 20\,Hz, together with robot proprioceptive state when supported by the method.

All policies predict 16-step action chunks and operate in the same control space, consisting of end-effector delta pose and gripper action. 
During evaluation, predicted action chunks are executed in an open-loop manner under the same control frequency and scene randomization protocol.

\subsection{Evaluation Metric}
Binary success is too coarse for long-horizon control, we therefore use a stage-wise scoring scheme.
For each task, we decompose the procedure into $N$ atomic sub-steps, defined as the smallest verifiable units of progress. Each completed sub-step receives $\frac{100}{N}$ points, and the episode score is
\[
\text{score} = \frac{\#\text{completed}}{N} \times 100.
\]

\textbf{Task-level aggregation.}
For each task, we evaluate 10 episodes with distinct initializations. 
The task score of a model is the mean episode score across these runs, and we also report the standard deviation.

\textbf{Mechanism-level aggregation.}
To evaluate capability- and ambiguity-specific performance, we average task scores over all tasks labeled with a given property. 
For Context-Independent tasks, capability scores are computed for Phase Dependence (PD), Iterative Progress (IP), Error Accumulation (EA), and Temporal Windows (TW) by averaging over the tasks associated with each structural property. 
For Context-Dependent tasks, ambiguity scores are computed analogously for Completion (CP), Count (CT), Subtask-Branch (SB), and Cross-Episode (CE).

\textbf{Regime-level aggregation.}
Context-Independent and Context-Dependent scores are computed by averaging task scores within each regime.

This multi-level evaluation protocol enables LongBench to measure not only overall performance, but also where performance degrades across different sources of long-horizon difficulty.

\subsection{Benchmark Access and Reproducibility}
\label{sec:reproducibility}

Real-world manipulation benchmarks are inherently harder to reproduce than simulation suites, as exact replication typically requires access to identical hardware, sensors, and scene assets. 
To make LongBench practically usable while preserving fair comparisons, we adopt a two-pronged reproducibility strategy: centralized evaluation and environment replication.

\textbf{Centralized evaluation and leaderboard.}
We provide a submission platform that allows users to evaluate their policies under a standardized setup without requiring local robot access. 
Submitters can either host their policy via a remote inference API or upload model checkpoints for evaluation. 
All submissions are evaluated under a fixed protocol with consistent initial conditions and environment settings, ensuring comparability across methods. 
Results are reported on a public leaderboard, and each submission receives detailed evaluation logs and optional rollout visualizations.

\textbf{Environment replication.}
To support local development and ablation studies, we will release a detailed object list with purchase links and a setup manual describing camera placement, calibration, scene initialization, and task definitions. 
This enables labs to reconstruct the benchmark environment, while the centralized leaderboard serves as a consistent reference for cross-method comparison.


\section{Benchmarking Results}
\label{sec:Benchmarking Results}

We use LongBench to diagnose how current policies fail under different sources of long-horizon difficulty. 
Rather than focusing on aggregate performance, our goal is to understand where performance breaks down across capability subsets, ambiguity types, and regimes.
Concretely, we investigate the following questions:

\begin{enumerate}[leftmargin=1.8em, labelsep=0.5em, itemsep=0.2em, topsep=0.2em]
    \item How do different policies perform across Context-Independent capability subsets, and which types of execution difficulty remain challenging?
    
    \item How does performance vary across Context-Dependent ambiguity patterns, and which forms of contextual dependency are not well handled by current methods?
    
    \item How do models behave across regimes, and do any policies achieve consistently strong performance on both Context-Independent and Context-Dependent tasks?
\end{enumerate}

\subsection{Results on Context-Independent Long-Horizon Tasks}

Context-Independent long-horizon tasks are \emph{fully observable}: the next action can be determined from the current observation without requiring hidden information from earlier steps. 
They isolate the ability to execute under extended rollouts, including stability under noise, contact-rich interactions, and feasibility constraints.

\begin{figure}[t]
  \centering
  \includegraphics[width=1.0\linewidth]{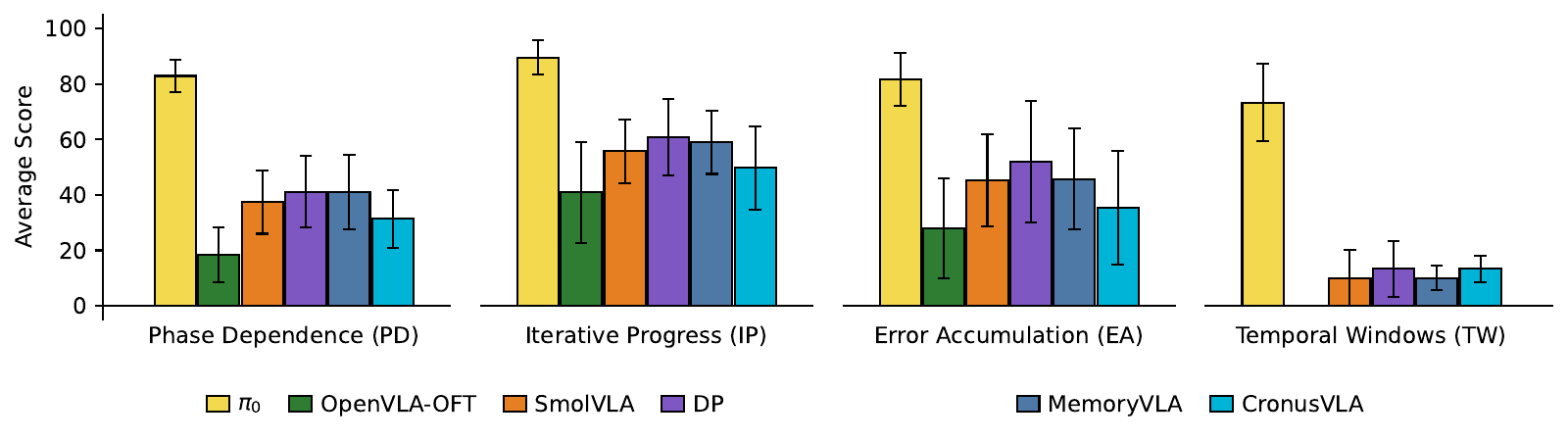}
  \caption{Capability-aligned performance on Context-Independent long-horizon tasks. We group the five Context-Independent long-horizon tasks by four structural capabilities—PD (Phase Dependence), IP (Iterative Progress), EA (Error Accumulation), and TW (Temporal Windows)—and report, for each capability, the mean success score of each policy over the tasks exhibiting that capability. Error bars denote the standard error of the mean (SEM) computed from all evaluation trials pooled across the tasks within each capability group.}
  \label{fig:lh_capability_bar}
\end{figure}

\textbf{Overview.}
Table~\ref{tab:ci_lh_results} summarizes task-level performance on Context-Independent long-horizon tasks. Even in fully observable settings, models exhibit large performance gaps.
As shown in Figure~\ref{fig:lh_capability_bar}, $\pi_0$ consistently outperforms all other methods across all subsets, while the remaining models form a substantially weaker second tier. 
This gap is not uniform: most methods retain partial competence on phase dependence, iterative progress, and error accumulation, but performance drops sharply on temporal-window tasks for all models except $\pi_0$. 

Beyond average performance, execution stability varies substantially. Error accumulation (EA) shows consistently larger variance across methods than other capability dimensions, indicating a general challenge in maintaining robustness under long-horizon error propagation. Even for $\pi_0$, the large error bars suggest that strong average performance does not imply consistent execution.

Overall, Context-Independent long-horizon performance is not governed by a single difficulty axis, but by uneven robustness across capability dimensions.

\begin{table}[t]
\centering
\small
\setlength{\tabcolsep}{4pt}
\caption{Performance comparison of 6 robotic manipulation policies on Context-Independent long-horizon tasks. Results are reported as mean stage-wise completion scores (\%).}
\label{tab:ci_lh_results}
\resizebox{\textwidth}{!}{%
\begin{tabular}{lcccccc}
\toprule
model & waste sorting & thread rope & pull drawer & stack block & dynamic grasping & average \\
\midrule
$\pi_0$ \citep{pi0}                     & 100.0 & 72.0 & 95.0 & 91.3 & 73.3 & 86.3 \\
OpenVLA-OFT \citep{kim2025fine}         & 90.0  & 46.0 & 17.5 & 10.0 & 0.0  & 32.7 \\
SmolVLA \citep{smolvla}                 & 83.3  & 62.0 & 48.8 & 28.8 & 10.0 & 46.6 \\
DP \citep{chi2023diffusionpolicy}       & 91.7  & 30.0 & 47.5 & 73.8 & 13.3 & 51.2 \\
MemoryVLA \citep{shi2026memoryvla}      & 81.6  & 64.0 & 62.5 & 27.5 & 10.0 & 49.1 \\
CronusVLA \citep{li2025cronusvla}       & 86.6  & 56.0 & 41.2 & 15.0 & 13.3 & 42.4 \\
\bottomrule
\end{tabular}%
}
\end{table}

\textbf{Consistent failure across execution-oriented capabilities.}
Across phase dependence, iterative progress, and error accumulation, models exhibit highly similar rankings despite differences in task structure. 
This consistency suggests that these subsets exhibit closely related performance patterns across current policies: while partial progress can be achieved, most methods struggle to maintain reliable performance over extended interaction sequences. 
As a result, errors introduced in early or intermediate stages frequently lead to incomplete execution, producing similar failure patterns across these capability subsets.

\textbf{Temporal-window tasks expose a critical weakness.}
In contrast to the gradual degradation observed in other subsets, performance on temporal-window tasks collapses for all models except $\pi_0$ ($73.3$ vs.\ $0.0$--$13.3$), indicating a sharp difference in performance across models. 
Notably, even methods that perform competitively on other Context-Independent long-horizon subsets fail almost completely in this setting. 
This consistent failure across models highlights a challenging aspect of current policy performance in this setting: they remain markedly less robust on the time-critical scenarios captured by temporal-window tasks.

\textbf{Implication.}
Our results suggest that full observability alone does not guarantee robust performance on Context-Independent long-horizon tasks. 
Rather than failing uniformly, current policies exhibit systematic weaknesses on specific capability subsets. 
In particular, while moderate performance can be achieved on standard execution-oriented tasks, most methods fail to handle more demanding scenarios such as temporal-window tasks. 
These results suggest that existing policies do not yet provide uniformly strong performance across all Context-Independent long-horizon capability subsets.

\begin{table}[t]
\centering
\small
\setlength{\tabcolsep}{4pt}
\caption{Performance comparison of 6 robotic manipulation policies on Context-Dependent long-horizon tasks. Results are reported as mean stage-wise completion scores (\%).}
\label{tab:cd_lh_results}
\resizebox{\textwidth}{!}{%
\begin{tabular}{lcccccc}
\toprule
model & vase sticks & wipe plate & repeat placement & swap blocks & hide block & average \\
\midrule
$\pi_0$ \citep{pi0}                     & 60.0 & 65.0 & 38.8 & 23.0 & 0.0 & 37.3 \\
OpenVLA-OFT \citep{kim2025fine}         & 56.6 & 2.5  & 23.8 & 0.0  & 0.0 & 16.6 \\
SmolVLA \citep{smolvla}                 & 48.3 & 43.8 & 43.8 & 22.0 & 55.0 & 42.6 \\
DP \citep{chi2023diffusionpolicy}       & 16.6 & 68.8 & 7.5  & 9.0  & 4.0 & 21.2 \\
MemoryVLA \citep{shi2026memoryvla}      & 56.6 & 82.5 & 27.5 & 36.0 & 24.0 & 45.3 \\
CronusVLA \citep{li2025cronusvla}       & 50.0 & 67.5 & 12.5 & 14.0 & 20.0 & 32.8 \\
\bottomrule
\end{tabular}%
}
\end{table}

\begin{figure}[t]
  \centering
  \includegraphics[width=1.0\linewidth]{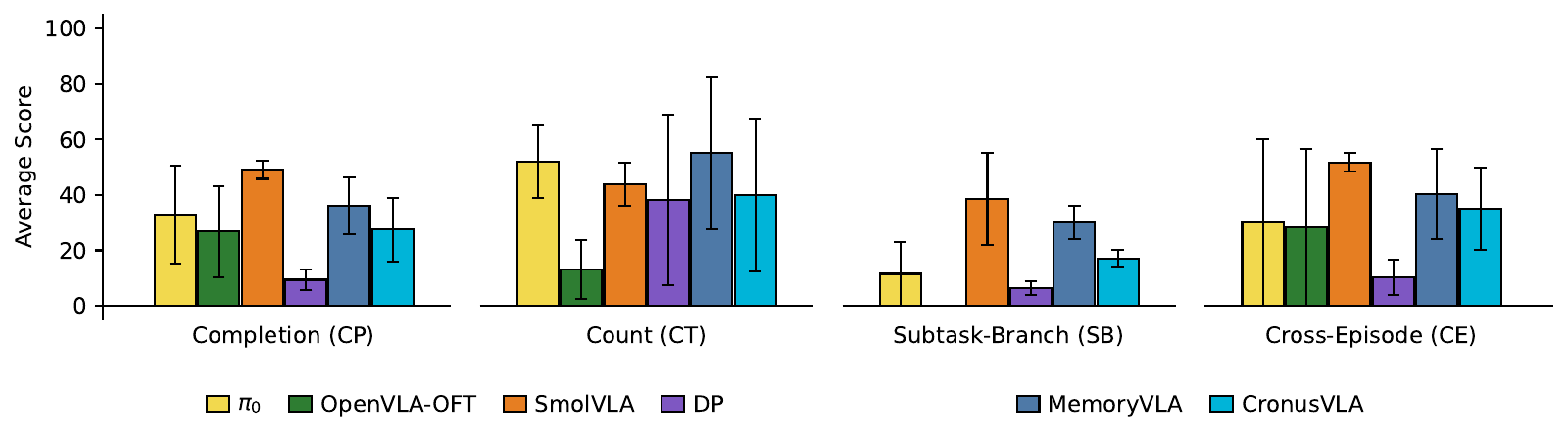}
  \caption{Ambiguity-pattern-aligned performance on Context-Dependent long-horizon tasks. The five Context-Dependent long-horizon tasks are grouped by four ambiguity patterns—CP (Completion), CT (Count), SB (Subtask-Branch), and CE (Cross-Episode). For each pattern, we report the average success score of each policy over the tasks associated with that pattern. Error bars show the standard error of the mean (SEM) computed over all evaluation episodes from all tasks in the corresponding pattern group.}

  \label{fig:lc_ambiguity_bar}
\end{figure}

\subsection{Results on Context-Dependent Long-Horizon Tasks}

Context-Dependent Long-Horizon tasks introduce ambiguity that cannot be resolved from the current observation alone. 
Visually similar states may correspond to different latent conditions, requiring policies to leverage historical context for correct action selection. 

\textbf{Overview.}
Table~\ref{tab:cd_lh_results} summarizes task-level performance on Context-Dependent long-horizon tasks. Tasks that appear to require similar forms of contextual reasoning exhibit drastically different difficulty. 
As shown in Figure~\ref{fig:lc_ambiguity_bar}, performance ranges from moderate success in count ambiguity (CT, up to 55.0) to near-complete failure in subtask-branch ambiguity (SB), where several models achieve near-zero scores. 
Completion (CP) and cross-episode (CE) ambiguity fall in between, but with large variation across policies. Notably, even in comparatively easier patterns such as CT and CE, performance exhibits substantial variance across policies, indicating poor stability and inconsistent utilization of contextual information.
This sharp disparity indicates that contextual tasks do not appear to follow a single unified difficulty axis, but instead exhibit substantially different difficulty characteristics despite their shared reliance on history.

\textbf{Memory helps, but does not generalize across ambiguity types.}
Memory-conditioned models achieve strong gains on count ambiguity (CT), with MemoryVLA reaching 55.0 and outperforming most baselines. 
However, similar improvements are not consistently observed on more structured ambiguity types. 
On completion (CP) and cross-episode (CE) tasks, memory-based models do not consistently outperform strong frame-based policies, and are sometimes surpassed by them (e.g., SmolVLA). 
This inconsistency suggests that access to history alone is not sufficient to consistently improve performance across all ambiguity types.

\textbf{Even with memory, policies struggle to leverage context consistently.}
Subtask-Branch ambiguity (SB) remains challenging across all methods, with near-zero performance for several models and only modest gains from memory (e.g., MemoryVLA: 30.0). 
This pattern suggests that retaining context alone may not be sufficient, and effective use of context remains challenging. 
Policies frequently produce actions inconsistent with the underlying subtask branch, indicating difficulty in using earlier context reliably.

\textbf{Implication.}
Our results suggest that access to history alone does not consistently lead to strong performance in context-dependent long-horizon tasks. 
While memory enables simple temporal reasoning (e.g., counting), effective performance is not guaranteed by access to history alone and varies across different forms of contextual dependency. 
These results suggest that context-dependent behavior is not a single memory problem, but involves multiple forms of contextual dependency that current policies fail to handle consistently.

\begin{figure}[t]
  \centering
  \includegraphics[width=0.72\linewidth]{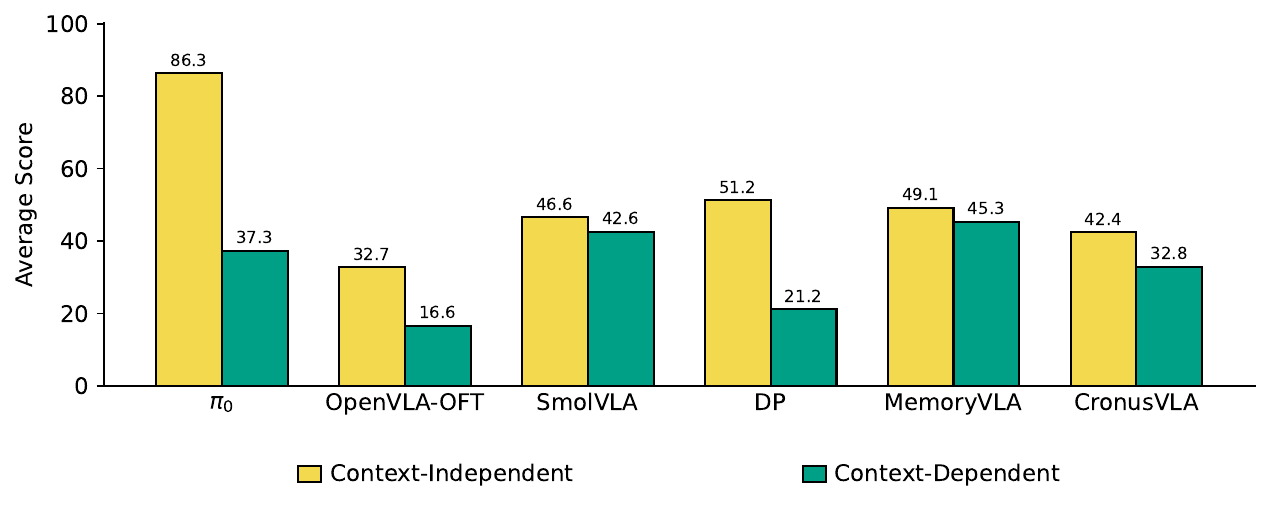}
  \caption{Regime-level performance of 6 manipulation policies on Context-Independent and Context-Dependent tasks. While Context-Dependent tasks induce consistent performance drops, models exhibit distinct cross-regime profiles: strong execution does not guarantee context robustness, and memory improves retention but does not uniformly improve overall performance. }
  \label{fig:lh_lc_bar}
\end{figure}

\subsection{Cross-Regime Analysis: Context-Independent vs. Context-Dependent Tasks}

We now compare performance across Context-Independent and Context-Dependent tasks to understand how different models generalize between fully observable and ambiguity-driven settings.

\textbf{Overview.}
Performance differs sharply across regimes. 
As shown in Figure~\ref{fig:lh_lc_bar}, $\pi_0$ achieves the highest Context-Independent performance (86.3) but exhibits the largest degradation under Context-Dependent tasks (37.3), while MemoryVLA shows the smallest performance gap (49.1 vs.\ 45.3) without achieving stronger overall performance.

\textbf{Distinct cross-regime performance profiles.}
Models that perform strongly in fully observable settings tend to degrade more under contextual ambiguity, while models with relatively stable performance under ambiguity do not achieve strong execution performance. 
This pattern indicates that different models exhibit different strengths across the two regimes.

\textbf{Explicit memory does not consistently lead to improved performance across regimes.}
SmolVLA, despite lacking explicit memory, achieves competitive performance on Context-Depen-dent tasks and outperforms several history-conditioned models. 
This result indicates that the presence of explicit memory alone does not ensure improved contextual performance. 
Instead, memory-based methods show uneven performance across different ambiguity types, suggesting that access to history does not consistently translate into better outcomes.

\textbf{Implication.}
The performance gap across regimes highlights that current models exhibit difficulty in achieving strong performance on both execution-oriented and context-dependent tasks simultaneously. 
Neither strong execution in fully observable settings nor the presence of explicit memory alone leads to robust performance across regimes. This suggests that long-horizon robustness cannot be captured by a single performance axis, but instead requires jointly reasoning about execution reliability and context-dependent decision making.
\section{Conclusion}
\label{sec:conclusion}

We introduce \textbf{LongBench}, a real-world benchmark for long-horizon robotic manipulation. 
It separates temporal difficulty into two complementary regimes: Context-Independent tasks, which emphasize sustained execution under full observability, and Context-Dependent tasks, which require resolving ambiguity using historical context. By organizing tasks into capability- and ambiguity-specific subsets, LongBench enables diagnostic evaluation beyond aggregate task success.

Using this benchmark, we observe that long-horizon performance exhibits clear non-uniform patterns. In Context-Independent tasks, performance varies across structural capability subsets, with temporal-window tasks consistently remaining challenging. In Context-Dependent tasks, performance differs substantially across ambiguity types, and improvements from memory-conditioned models are not consistently observed across all cases. Across regimes, no evaluated policy achieves uniformly strong performance on both execution-oriented and context-dependent tasks.

Our analysis is diagnostic rather than causal: we do not attribute these patterns to specific internal mechanisms of individual models. Instead, we hope LongBench provides a structured evaluation framework for studying long-horizon manipulation and for developing methods that are robust across multiple sources of temporal difficulty.







{\small
\bibliographystyle{apalike}
\bibliography{sample-base}

@String{Computer = "{IEEE} Computer" }

@String{Chelsea = "Chelsea" }

@article{rlbench,
  title={RLBench: The Robot Learning Benchmark and Learning Environment},
  author={Stephen James and Zicong Ma and David Rovick Arrojo and Andrew J. Davison},
  journal={arXiv preprint},
  year={2020},
  url={https://sites.google.com/view/rlbench}
}

@article{calvin,
  title={CALVIN: A Benchmark for Language-Conditioned Policy Learning for Long-Horizon Robot Manipulation Tasks},
  author={Oier Mees and Lukas Hermann and Erick Rosete-Beas and Wolfram Burgard},
  journal={IEEE Robotics and Automation Letters},
  year={2022},
  url={https://arxiv.org/abs/2112.03227}
}

@article{libero,
  title={LIBERO: Benchmarking Knowledge Transfer for Lifelong Robot Learning},
  author={Bo Liu and Yifeng Zhu and Chongkai Gao and Yihao Feng and Qiang Liu and Yuke Zhu and Peter Stone},
  journal={Advances in Neural Information Processing Systems},
  year={2023},
  url={https://openreview.net/forum?id=xzEtNSuDJk}
}

@inproceedings{vlabench,
  title={VLABench: A Large-Scale Benchmark for Language-Conditioned Robotics Manipulation with Long-Horizon Reasoning Tasks},
  author={Shiduo Zhang and Zhe Xu and Peiju Liu and Xiaopeng Yu and Yuan Li and Qinghui Gao and Zhaoye Fei and Zhangyue Yin and Zuxuan Wu and Yu-Gang Jiang and Xipeng Qiu},
  booktitle={Proceedings of the IEEE/CVF International Conference on Computer Vision},
  year={2025},
  url={https://openaccess.thecvf.com/content/ICCV2025/papers/Zhang_VLABench_A_Large-Scale_Benchmark_for_Language-Conditioned_Robotics_Manipulation_with_Long-Horizon_ICCV_2025_paper.pdf}
}

@article{robocerebra,
  title={RoboCerebra: A Large-Scale Benchmark for Long-Horizon Robotic Manipulation Evaluation},
  author={Songhao Han and Boxiang Qiu and Yue Liao and Siyuan Huang and Chen Gao and Shuicheng Yan and Si Liu},
  journal={arXiv preprint},
  year={2025},
  url={https://arxiv.org/html/2506.06677v1}
}

@article{robocas,
  title={RoboCAS: A Benchmark for Robotic Manipulation in Complex Object Arrangement Scenarios},
  author={Liming Zheng and Feng Yan and Fanfan Liu and Chengjian Feng and Zhuoliang Kang and Lin Ma},
  journal={arXiv preprint},
  year={2024},
  url={https://arxiv.org/abs/2407.06951}
}

@article{kim2025fine,
  title={Fine-Tuning Vision-Language-Action Models: Optimizing Speed and Success},
  author={Kim, Moo Jin and Finn, Chelsea and Liang, Percy},
  journal={arXiv preprint arXiv:2502.19645},
  year={2025}
}

@inproceedings{chi2023diffusionpolicy,
	title={Diffusion Policy: Visuomotor Policy Learning via Action Diffusion},
	author={Chi, Cheng and Feng, Siyuan and Du, Yilun and Xu, Zhenjia and Cousineau, Eric and Burchfiel, Benjamin and Song, Shuran},
	booktitle={Proceedings of Robotics: Science and Systems (RSS)},
	year={2023}
}

@article{smolvla,
  title         = {SmolVLA: A Vision-Language-Action Model for Affordable and Efficient Robotics},
  author        = {Shukor, Mustafa and Aubakirova, Dana and Capuano, Francesco and Kooijmans, Pepijn and Palma, Steven and Zouitine, Adil and Aractingi, Michel and Pascal, Caroline and Russi, Martino and Marafioti, Andres and Alibert, Simon and Cord, Matthieu and Wolf, Thomas and Cadene, Remi},
  journal       = {arXiv preprint arXiv:2506.01844},
  year          = {2025}
}

@article{pi0,
  title         = {$\pi_0$: A Vision-Language-Action Flow Model for General Robot Control},
  author        = {Black, Kevin and Brown, Noah and Driess, Danny and Esmail, Adnan and Equi, Michael and Finn, Chelsea and Fusai, Niccolo and Groom, Lachy and Hausman, Karol and Ichter, Brian and Jakubczak, Szymon and Jones, Tim and Ke, Liyiming and Levine, Sergey and Li-Bell, Adrian and Mothukuri, Mohith and Nair, Suraj and Pertsch, Karl and Shi, Lucy Xiaoyang and Tanner, James and Vuong, Quan and Walling, Anna and Wang, Haohuan and Zhilinsky, Ury},
  journal       = {arXiv preprint arXiv:2410.24164},
  year          = {2024}
}

@article{heo2023furniturebench,
  title   = {FurnitureBench: Reproducible Real-World Benchmark for Long-Horizon Complex Manipulation},
  author  = {Heo, Minho and Lee, Youngwoon and Lee, Doohyun and Lim, Joseph J.},
  journal = {Robotics: Science and Systems (RSS)},
  year    = {2023},
  url     = {https://arxiv.org/abs/2305.12821}
}

@article{khargonkar2024scenereplica,
  title   = {SCENEREPLICA: Benchmarking Real-World Robot Manipulation by Creating Replicable Scenes},
  author  = {Khargonkar, Ninad and Allu, Sai Haneesh and Lu, Yangxiao and Jaykumar P, Jishnu and Prabhakaran, Balakrishnan and Xiang, Yu},
  journal = {IEEE International Conference on Robotics and Automation (ICRA)},
  year    = {2024},
  url     = {https://arxiv.org/abs/2306.15620}
}

@misc{rgmc2025,
  title  = {Robotic Grasping and Manipulation Competition (RGMC): Manufacturing Track},
  author = {{National Institute of Standards and Technology}},
  year   = {2023},
  url    = {https://www.nist.gov/el/intelligent-systems-division-73500/robotic-grasping-and-manipulation-assembly/robotic-grasping}
}

@article{atreya2025roboarena,
  title   = {RoboArena: Distributed Real-World Evaluation of Generalist Robot Policies},
  author  = {Atreya, Pranav and Pertsch, Karl and Lee, Tony and Kim, Moo Jin and Jain, Arhan and others},
  journal = {arXiv preprint arXiv:2506.18123},
  year    = {2025},
  url     = {https://arxiv.org/abs/2506.18123}
}

@article{zhou2025autoeval,
  title   = {AutoEval: Autonomous Evaluation of Generalist Robot Manipulation Policies in the Real World},
  author  = {Zhou, Zhiyuan and Atreya, Pranav and Tan, You Liang and Pertsch, Karl and Levine, Sergey},
  journal = {arXiv preprint arXiv:2503.24278},
  year    = {2025},
  url     = {https://arxiv.org/abs/2503.24278}
}

@misc{shi2026memoryvla,
      title={MemoryVLA: Perceptual-Cognitive Memory in Vision-Language-Action Models for Robotic Manipulation}, 
      author={Hao Shi and Bin Xie and Yingfei Liu and Lin Sun and Fengrong Liu and Tiancai Wang and Erjin Zhou and Haoqiang Fan and Xiangyu Zhang and Gao Huang},
      year={2026},
      eprint={2508.19236},
      archivePrefix={arXiv},
      primaryClass={cs.RO},
      url={https://arxiv.org/abs/2508.19236}, 
}

@misc{li2025cronusvla,
      title={CronusVLA: Towards Efficient and Robust Manipulation via Multi-Frame Vision-Language-Action Modeling}, 
      author={Hao Li and Shuai Yang and Yilun Chen and Xinyi Chen and Xiaoda Yang and Yang Tian and Hanqing Wang and Tai Wang and Dahua Lin and Feng Zhao and Jiangmiao Pang},
      year={2025},
      eprint={2506.19816},
      archivePrefix={arXiv},
      primaryClass={cs.RO},
      url={https://arxiv.org/abs/2506.19816}, 
}
}

\clearpage
\appendix
\section*{Appendix}
\section{Implementation Details}
\label{app:more_results}

For clarity, we summarize the training protocols of all evaluated policies in Table~\ref{tab:impl_details}. All methods are trained on the same LongBench demonstrations in LeRobot format. Across all policies, we adopt an action chunk size of \textbf{16} and evaluate by open-loop execution of the predicted action chunk. We additionally report whether a method uses observation history, as this is a core modeling distinction between single-frame and multi-frame policies: $\pi_0$, OpenVLA-OFT, SmolVLA, and Diffusion Policy operate on single-frame observations, whereas MemoryVLA and CronusVLA incorporate temporal history. All other method-specific choices, such as architecture details, internal resizing and preprocessing, proprioceptive inputs, optimizer settings, training schedules, and hardware configurations, follow the original papers and official implementations.

\section{Additional Visualizations}
\label{app:qual_examples}

\begin{figure}[t]
  \centering
  \includegraphics[width=0.9\linewidth]{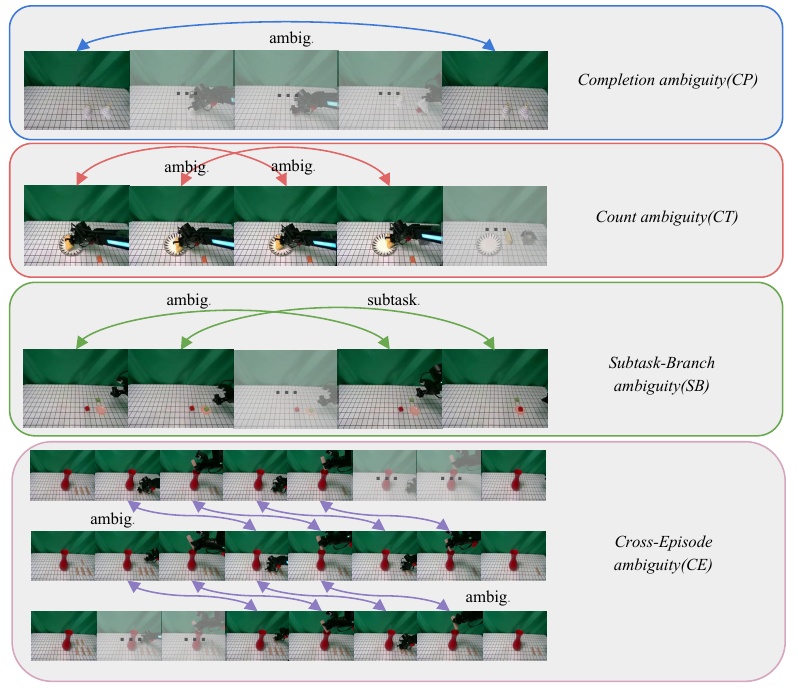}
  \caption{Representative ambiguity mechanisms in Context-Dependent tasks. We visualize four capability types: completion ambiguity (CP), count ambiguity (CT), subtask-branch ambiguity (SB), and cross-episode ambiguity (CE). Arrows highlight ambiguous states where the correct action cannot be determined from the current observation alone.}
  \label{fig:cdlh-steps}
\end{figure}

To better characterize Context-Dependent reasoning, we group tasks by the type of ambiguity they induce, as illustrated in {Figure~\ref{fig:cdlh-steps}}.
\textbf{Completion ambiguity (CP)} arises when two observations appear similar, but the correct action differs depending on whether a previously required step has already been completed. 
\textbf{Count ambiguity (CT)} occurs when the policy must remember how many times an action has been executed, since the same local observation may require either continuing or stopping depending on the accumulated count. 
\textbf{Subtask-branch ambiguity (SB)} appears when the current scene alone is insufficient to determine which branch of a multi-stage task the agent is currently following, making action selection depend on earlier subtask history. 
\textbf{Cross-episode ambiguity (CE)} is the most demanding case, where visually similar states recur across repeated phases of a trajectory, and the policy must retrieve information from much earlier context to choose the correct behavior.

\begin{figure}[t]
  \centering
  \includegraphics[width=\textwidth]{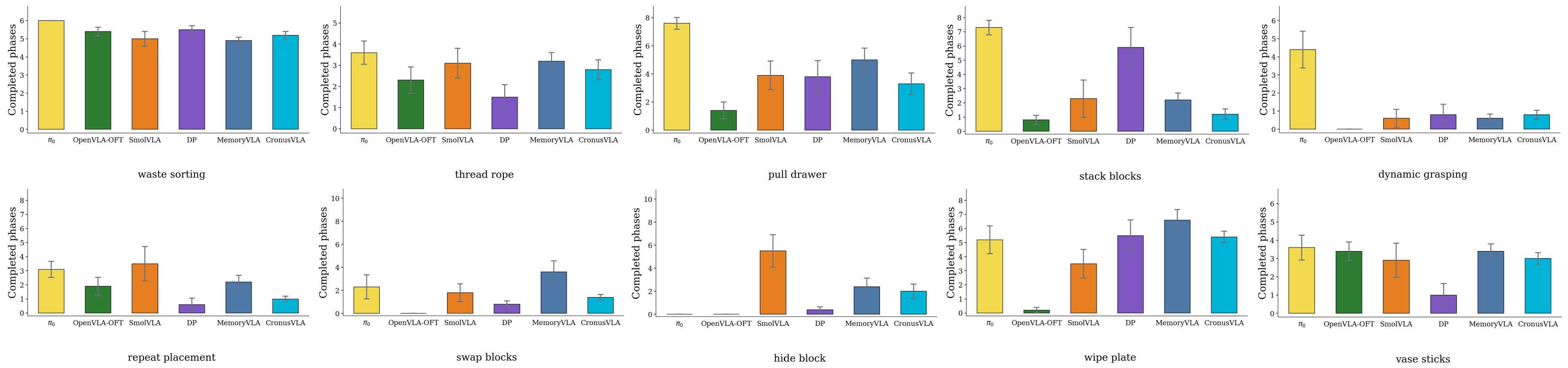}
  \caption{Completed-phase statistics across the ten tasks. Bars denote the mean number of completed phases over 10 episodes for each model, and error bars indicate the standard error of the mean (SEM). Context-Independent tasks are shown in the top row, and Context-Dependent tasks are shown in the bottom row.}
  \label{fig:stage-counts}
\end{figure}

Fig.~\ref{fig:stage-counts} shows the average number of phases each policy completes before failure across the ten LongBench tasks. One counter-intuitive pattern is that multi-frame policies do not always fail later than single-frame policies. Under the usual expectation, models with temporal history should be more likely to progress through the early stages and fail only in later phases due to accumulated errors. However, this is not what we observe in several context-dependent tasks. On tasks such as \textit{repeat placement} and \textit{hide block}, MemoryVLA and CronusVLA do not consistently exhibit the expected trend of surviving longer. In many cases, they complete only a small number of phases, indicating that they are not simply breaking down near the end of execution, but instead making incorrect contextual judgments already in the early, decision-critical stages. In other words, the issue is not merely that these models fail to retain enough past information, but that they may retain the wrong cues, or commit too early to an incorrect hidden-state interpretation based on history. Once this happens, the rollout proceeds along the wrong branch from the outset. This suggests that the main challenge of context-dependent manipulation is not simply extending the temporal window, but correctly binding, retrieving, and resetting historical information in relation to the current visual evidence.

\begin{figure}[t]
  \centering
  \includegraphics[width=0.98\linewidth]{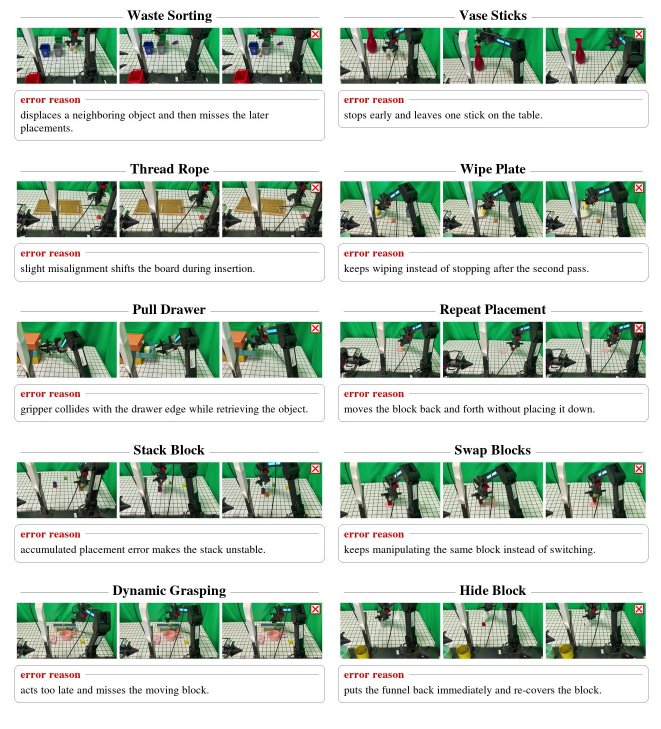}
  \caption{Representative failure cases across the ten tasks. The left column shows Context-Independent long-horizon tasks, while the right column shows Context-Dependent long-context tasks. For each task, we present three key frames in temporal order and mark the failed final state with a red cross; the box below gives a concise description of the immediate failure.}
  \label{fig:failure_sequences}
\end{figure}

Figure~\ref{fig:failure_sequences} summarizes representative failures across all ten tasks under the grouped visualization used in our benchmark. The left column highlights Context-Independent long-horizon failures, which are mainly associated with execution robustness, such as unintended object disturbance, contact misalignment, accumulated placement error, or missing a narrow temporal interaction window. In contrast, the right column highlights Context-Dependent long-context failures, where the dominant difficulty is ambiguity: visually similar intermediate observations can correspond to different latent task histories, causing the policy to stop too early, repeat an earlier action, or manipulate the wrong object. This side-by-side organization makes the distinction between robustness-driven long-horizon failures and memory-dependent long-context failures more explicit.
\begin{table}[t]
\centering
\small
\setlength{\tabcolsep}{3pt}
\renewcommand{\arraystretch}{1.05}
\caption{Training setup summary for all evaluated policies, including shared data settings (raw resolution, chunk size, and frequency) and key method-specific configurations (learning rate, batch size, and use of observation history). All other implementation details follow the original papers and official implementations.}
\label{tab:impl_details}
\resizebox{0.9\textwidth}{!}{
\begin{tabular}{lcccccc}
\toprule
\textbf{Detail} & $\boldsymbol{\pi_0}$ & \textbf{OpenVLA-OFT} & \textbf{SmolVLA} & \textbf{Diffusion Policy} & \textbf{MemoryVLA} & \textbf{CronusVLA} \\
\midrule
input resolution    & 224$\times$224 & 224$\times$224 & 512$\times$512 & 288$\times$216 & 224$\times$224 & 224$\times$224 \\
chunk size        & 16 & 16 & 16 & 16 & 16 & 16 \\
frequency(hz)     & 20 & 20 & 20 & 20 & 20 & 20 \\
learning rate     & 5e-4 & 5e-4 & 2e-4 & 1e-4 & 5e-6 & 2e-5 \\
batch size        & 32 & 64 & 64 & 64 & 64 & 32 \\
use history       & No & No & No & No & Yes & Yes \\
\bottomrule
\end{tabular}
}
\end{table}

\section{LongBench Tasks Description}
\label{app:task_description}

\begin{figure}[t]
  \centering
  \includegraphics[width=0.9\linewidth]{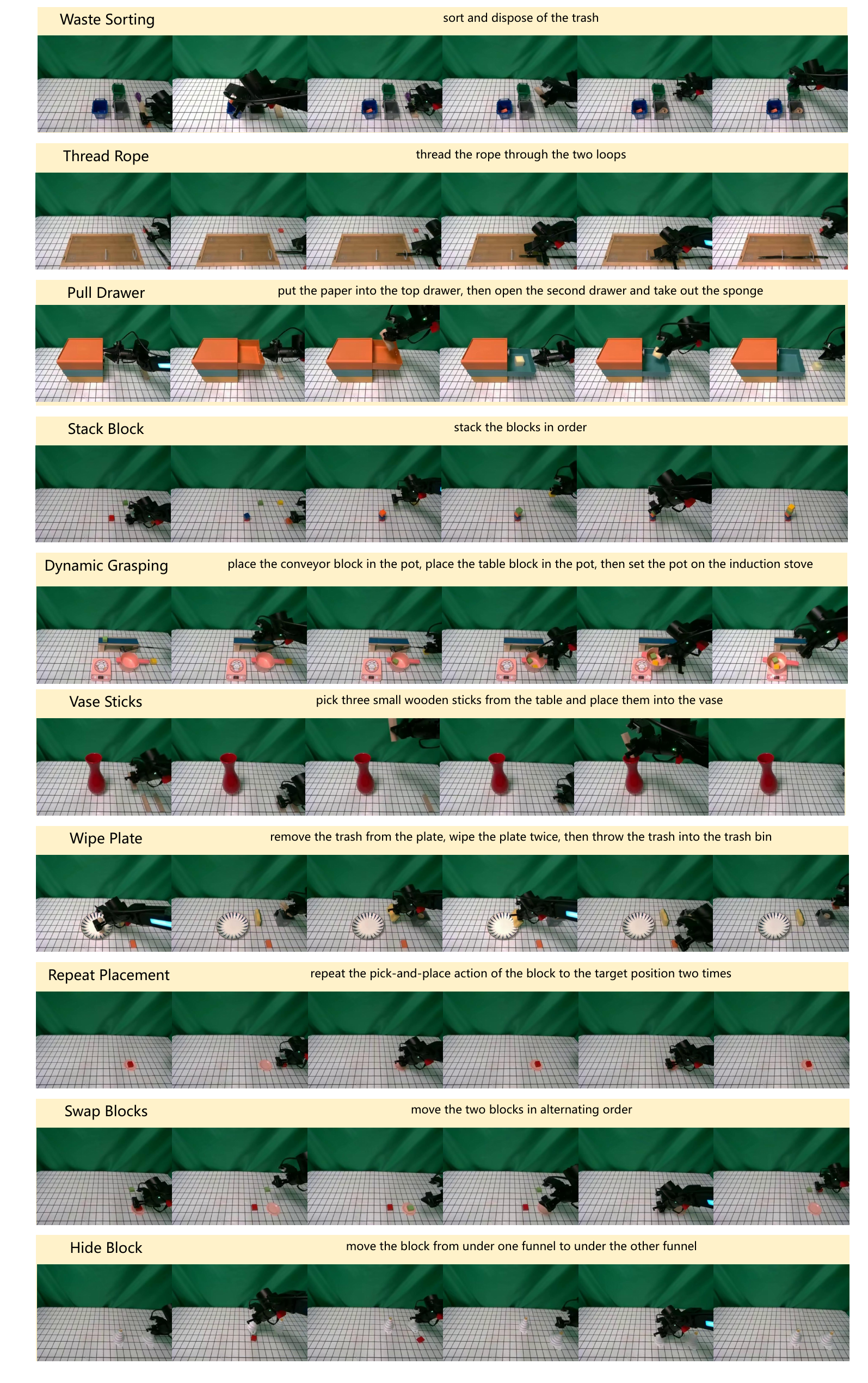}
 \caption{Representative rollout visualizations for the 10 LongBench tasks. Each row shows chronologically ordered key frames from one task trajectory.}
  \label{fig:rollout}
\end{figure}

\textbf{Task rollout visualizations.}
Figure~\ref{fig:rollout} presents rollout visualizations for all ten tasks in LongBench.
Each row shows a representative trajectory as a sequence of key frames sampled in chronological order, illustrating how the scene evolves throughout task execution.
The first five rows correspond to Context-Independent Long-Horizon tasks (\textit{Waste Sorting}, \textit{Thread Rope}, \textit{Pull Drawer}, \textit{Stack Block}, and \textit{Dynamic Grasping}), which require sustained multi-step manipulation over extended horizons.
The last five rows correspond to Context-Dependent Long-Horizon tasks (\textit{Vase Sticks}, \textit{Wipe Plate}, \textit{Repeat Placement}, \textit{Swap Blocks}, and \textit{Hide Block}), where correct actions depend on resolving ambiguity using context accumulated over time.
These visualizations provide an intuitive view of the temporal structure, interaction patterns, and contextual dependencies of the benchmark tasks.

\begin{table}[t]
\caption{Task descriptions of the LongBench benchmark.}
\label{tab:task_description}
\centering
\small
\renewcommand{\arraystretch}{1.5}
\setlength{\tabcolsep}{5pt}
\begin{tabularx}{\textwidth}{
>{\centering\arraybackslash\itshape}m{0.2\textwidth}
>{\raggedright\arraybackslash}X}
\toprule
\textbf{Task} & \textbf{Description} \\
\midrule

\makecell[c]{Waste Sorting} &
Multiple objects from different categories are placed on the table together with several target regions. The robot must identify each object category and execute a sequence of pick-and-place actions until all objects are sorted into the correct target areas. This task evaluates Context-Independent long-horizon sequencing and robustness under repeated manipulation. \\
\addlinespace
\hline

\makecell[c]{Thread Rope} &
A rope and two target holes are placed in the workspace. The robot must repeatedly adjust the rope pose, align the rope end with the openings, and continue corrective motions until the rope is successfully threaded through the holes. This task requires iterative progress and recovery from small alignment errors. \\
\addlinespace
\hline

Pull Drawer &
A drawer is initially closed, with task-relevant objects located inside or around it. The robot must first open the top drawer, place the object from the tabletop into the top drawer, and then close it. Next, the robot opens the middle drawer, takes out the object inside, places it on the tabletop, and finally closes the middle drawer. This task evaluates multi-step execution with strong phase dependence. \\
\addlinespace
\hline

Stack Block &
Multiple blocks are scattered on the table. The robot must sequentially pick up the blocks and place them into a stable stack, requiring accurate alignment and consistency across repeated steps. This task emphasizes cumulative error control over long manipulation sequences. \\
\addlinespace
\hline

Dynamic Grasping &
A target object moves along a conveyor belt during execution. The robot must track and grasp it within a limited temporal window before it moves out of reach, then continue the remaining manipulation by transferring and placing it at the target location. This task evaluates responsiveness and robustness under dynamic scene changes, as well as sustained execution after time-sensitive grasping.\\
\addlinespace
\hline

Vase Sticks &
Three to five thin sticks and a vase-like container are placed in the workspace. The robot must understand the instruction and insert exactly three sticks into the vase in sequence, neither more nor fewer. This task evaluates Context-Dependent understanding under cross-episode ambiguity, where successful execution depends on retaining and applying context across episodes. \\
\addlinespace
\hline

Wipe Plate &
A plate, a wiping tool, and multiple possible target regions or objects are placed on the table. Following the instruction, the robot must first discard the object on the plate into the trash bin, then wipe the plate exactly twice, neither more nor fewer, and finally discard the object on the table into the trash bin. This task tests whether the policy can retain and execute long-horizon instructions with precise temporal counting and stage-dependent action ordering. \\
\addlinespace
\hline

Repeat Placement &
A plate and a block are placed on the table. The robot is instructed to repeatedly perform a pick-and-place action, moving the block onto the plate exactly twice. Successful execution requires remembering the repetition count and maintaining consistent manipulation across steps. This task evaluates temporal counting and context retention over repeated actions. \\
\addlinespace
\hline

Swap Blocks &
Two blocks and a plate are placed on the table. The robot must repeatedly and in order place the two blocks onto the plate, and then return each of them to its original position once. This task tests whether the policy can preserve sequential context and correctly execute branching subtasks in a repeated multi-step manipulation process. \\
\addlinespace
\hline

Hide Block &
Initially, two funnels and one block are placed on the table, with the block hidden under one funnel. The robot must move the funnels aside, place the block under the other funnel, and return it so that the final scene looks identical to the initial one. This task evaluates Context-Dependent understanding under cross-episode ambiguity, requiring the policy to preserve hidden state beyond the visible scene. \\
\addlinespace
\hline

\end{tabularx}
\end{table}

\newpage

\end{document}